# Comparative Performance of Advanced NLP Models and LLMs in Multilingual Geo-Entity Detection

Kalin K. Kopanov

Modeling and Optimization Department, Institute of Information and Communication Technologies – Bulgarian Academy of Science, Sofia, Bulgaria, kalin.kopanov@iict.bas.bg

The integration of advanced Natural Language Processing (NLP) methodologies and Large Language Models (LLMs) has significantly enhanced the extraction and analysis of geospatial data from multilingual texts, impacting sectors such as national and international security. This paper presents a comprehensive evaluation of leading NLP models—SpaCy, XLM-RoBERTa, mLUKE, GeoLM—and LLMs, specifically OpenAI's GPT 3.5 and GPT 4, within the context of multilingual geo-entity detection. Utilizing datasets from Telegram channels in English, Russian, and Arabic, we examine the performance of these models through metrics such as accuracy, precision, recall, and F1 scores, to assess their effectiveness in accurately identifying geospatial references. The analysis exposes each model's distinct advantages and challenges, underscoring the complexities involved in achieving precise geo-entity identification across varied linguistic landscapes. The conclusions drawn from this experiment aim to direct the enhancement and creation of more advanced and inclusive NLP tools, thus advancing the field of geospatial analysis and its application to global security.

**CCS CONCEPTS** • Computing methodologies~Artificial intelligence~Natural language processing~Information extraction • Information systems~Information retrieval~Evaluation of retrieval results • Security and privacy~Human and societal aspects of security and privacy~Usability in security and privacy

## 1  Introduction

In recent years, the field of Artificial Intelligence (AI), with Natural Language Processing (NLP) as a pivotal branch, has seen a surge of innovation, driven by the emergence of cutting-edge models and computational techniques. These advancements have profoundly transformed our capacity to parse, comprehend, and infer from human language, marking a significant leap forward in both academic research and practical applications. A notable advance is the application of NLP in geospatial analysis, particularly geoparsing and event detection, which underscores the utility of textual data in extracting location-specific information.

The introduction of transformer-based architectures, notably BERT and its derivatives, has been a cornerstone in NLP's evolution. These models leverage deep learning to process large datasets, capturing subtle nuances of language that were previously inaccessible. For multilingual NLP, models such as XLM-RoBERTa represent a significant milestone, facilitating a unified approach to processing multiple languages, reducing the need for language-specific resources, and facilitating more inclusive and global analysis of textual data.

Models like mLUKE and GeoLM extend NLP's capabilities by embedding entity and geospatial knowledge directly into language processing. Nevertheless, it is still arguable how these models perform against other advanced models like XLM-RoBERTa in terms of efficiency in location-specific tasks. This comparison underscores the ongoing discussion about the optimal integration of specific domain knowledge, such as geographical information, within the NLP field.

The integration of Large Language Models (LLMs) like OpenAI's GPT-3.5 and GPT-4 into workflows significantly enhances NLP's ability to detect geospatial references, showcasing sophisticated context understanding. These models can discern subtle cues that indicate location, making them invaluable for tasks such as monitoring real-time events or analyzing data from social media platforms.

This paper presents a pivotal case study that evaluates the performance of advanced NLP models and LLMs in identifying locations within multilingual Telegram data related to ongoing conflicts, such as those in the Middle East and the Russian-Ukrainian Conflict. By examining content in English, Russian, and Arabic, this research assesses the efficacy of current technologies in multilingual environments and explores their potential contributions to crisis management, humanitarian response, and situational awareness.

Furthermore, this examination of conflict-based content underscores the critical importance of advanced NLP models in the field of national and international security. These technologies enhance our understanding and monitoring capabilities, holding the potential to inform and refine security monitoring, concerns, and actions on a global scale. Our goal is to assess these models based on accuracy and F-1 scores, providing a comprehensive analysis of their capabilities in geospatial and event-related information extraction from diverse linguistic sources.

## 2  Related work

Historically, Named Entity Recognition (NER) for locations relied on rule-based systems that utilized external sources, such as dictionaries and gazetteers [1]. These systems struggled with language ambiguity and lacked scalability, particularly in multilingual contexts. The shift towards advanced NLP models, exemplified by SpaCy's hybrid method that merges rules with machine learning, has significantly enhanced NER's efficiency and scalability across languages.

Advancements in pretrained language models (PLMs), including XLM-RoBERTa [2], LUKE [3], and mLUKE [4], have enabled cross-lingual learning and entity-aware processing, thereby improving the precision and context sensitivity essential for NER tasks. Furthermore, the debut of GeoLM [5] has propelled NLP's ability to recognize geospatial entities by using geographically enriched training datasets. Likewise, the emergence of GPT models, such as OpenAI's GPT 3.5 and GPT 4, has broadened NLP's scope in text generation and comprehension, facilitating the detection of sophisticated geo-entities among other applications.

## 3 Models and Data

A critical component of our research involves selecting high-performing NLP models and LLMs that can provide robust and accurate entity detection. In pursuit of this goal, we have chosen to include a diverse array of models in our comparative study: SpaCy, XLM-RoBERTa, mLUKE, GeoLM, OpenAI's GPT 3.5, and GPT 4. These models have been selected for their advanced capabilities and proven effectiveness in various NLP tasks, including but not limited to entity detection across multiple languages.

### 3.1 SpaCy

Among the numerous options available, SpaCy, a leading open-source library for advanced NLP, stands out as an excellent choice for several reasons. SpaCy is renowned for its speed and efficiency, making it highly suitable for large-scale NLP tasks. Furthermore, it offers a wide range of pre-trained models tailored for different languages [6], which are essential for our comparative study. The library's emphasis on providing models optimized for real-world applications aligns perfectly with our research objectives.

For our study, we selected the following SpaCy language models: English (en_core_web_md, v3.7.1), Russian (ru_core_news_md, v3.7.0), and Multi-language (xx_ent_wiki_sm, v3.7.0). The latter is specifically designed for processing unsupported languages, such as Arabic (presumably), with an exclusive focus on NER.

### 3.2 XLM-RoBERTa

In our geo-entity detection research, we employ the "xlm-roberta-large-finetuned-conll03-english" [7] model for its NER capabilities in English, derived from fine-tuning on the CoNLL-2003 dataset. This process leverages the model's extensive pretraining on the 2.5TB CommonCrawl corpus, ensuring precise entity identification within English texts.

For analyzing texts in other languages, including Russian and Arabic, the study continues to utilize the latter model. Despite the fine-tuning on English NER tasks, the model's underlying performance for non-English languages is supported by the broad linguistic knowledge embedded in the original "xlm-roberta-large" architecture. This extensive pretraining provides a robust foundation for accurate geo-entity recognition across languages, allowing the model to apply its generalized cross-lingual capabilities to a wide range of linguistic contexts.

### 3.3 mLUKE

Our methodology integrates mLUKE with SpaCy to enhance NER by leveraging SpaCy for initial entity detection and character span identification in unstructured texts. This integration allows for precise entity extraction, crucial for mLUKE's advanced entity-centric processing across languages. mLUKE, building on LUKE's principles and incorporating XLM-RoBERTa's cross-lingual capabilities, extends its application to a broader linguistic scope. The process begins with language detection, where SpaCy's targeted models for detected language undertake efficient NER tasks. Subsequently, the character spans detected by SpaCy are aligned with mLUKE's required token spans, using the "studio-ousia/mluke-large-lite-finetuned-conll-2003" model [8], facilitating a seamless transition to in-depth, language-specific analysis.

### 3.4 GeoLM

The GeoLM model plays a critical role in the field of geospatially informed NLP by advancing the accuracy of toponym recognition. Pre-trained on a rich amalgamation of datasets including OpenStreetMap (OSM), WikiData, and Wikipedia, and subsequently fine-tuned on the GeoWebNews dataset, the model is instantiated through the "zekun-li/geolm-base-toponym-recognition" implementation [9]. Tailored specifically for the English language, this model demonstrates exceptional proficiency in identifying and classifying geographical names within English narratives, thereby showcasing its capability to process geospatial information embedded in natural language effectively.

However, the model's English-centric optimization does pose limitations on its applicability to texts in other languages, attributable to the distinct linguistic structures, toponymic conventions, and geospatial nuances inherent to each language. These constraints underscore the need for adaptation or extension of the GeoLM model to ensure its utility across a broader linguistic spectrum.

### 3.5 OpenAI's GPT 3.5 and GPT 4

Our study utilizes OpenAI's LLM, GPT-3.5 and GPT-4, to automate the recognition of geographical entities in NER tasks. These models excel at identifying and classifying geographical names—such as cities, countries, and landmarks—across multiple languages, enhancing the extraction of location-based information from text. Their multilingual capabilities are particularly advantageous for geographical NER, allowing for broad application in global contexts without the need for language-specific optimizations.

For empirical validation, the experiment employs "gpt-3.5-turbo-0125" and "gpt-4-0125-preview" models accessed via API [10]. A custom prompt directs these models to focus on geographical locations, ensuring output precision by filtering out irrelevant elements. This approach demonstrates the models' ability to accurately process and extract geographical entities, highlighting their significance in advancing NLP's role in geospatial analysis.

### 3.6 Data

To evaluate the performance of NLP models and LLMs in detecting geo-entities across multiple languages, we curated a dataset from diverse, multilingual Telegram channels, focusing on January and February 2024. This dataset encompasses geopolitical discourse, news aggregation, and global event analysis in English, Russian, and Arabic, providing a comprehensive linguistic and cultural spectrum.

Our dataset includes posts from the following languages: for English, we extracted 840 posts from Intel Slava Z [11], a Russian news aggregator focusing on global conflicts and geopolitics, and 214 posts from The Rage X [12], which analyzes global events. The Russian segment comprises 2,406 posts from Два майора (Dva Mayora) [13], providing extensive content for analysis. For Arabic, 1,065 posts were sourced from المركز الإعلامي الحديدة (ALHodeidah Media Center) [14], the media wing of Ansar Allah in Yemen, offering continuous coverage of local developments. This diverse compilation from Telegram channels, spanning English, Russian, and Arabic, supplies a rich corpus for exploring linguistic and semantic nuances in geo-entity detection, crucial for understanding global conflicts and their impacts through media coverage.

It's essential to note that the channels selected for this study serve merely as a real-world sample and are not intended to measure their bias, reach or importance. Instead, they are used to visualize the importance of media coverage in creating awareness about conflict situations, showcasing the potential of advanced NLP and LLM technologies in processing and interpreting multilingual data for geopolitical analysis.

## 4 EVALUATION

The empirical validation of our research is anchored in a detailed evaluation framework, utilizing a custom Python script to analyze text data extracted from identified Telegram channels. This script leverages a suite of models—SpaCy, XLM-RoBERTa, mLUKE, GeoLM, OpenAI's GPT 3.5, and GPT 4—to parse multilingual content and identify geographical entities. These models were chosen for their proven efficacy in NLP tasks, ability to process multiple languages, and innovative entity detection methodologies.

### 4.1 Data Processing

Initially, the script preprocesses the raw data by applying normalization techniques. This step is crucial for mitigating variations in formatting and encoding across languages, ensuring that all models operate on a uniform dataset. Subsequently, the script employs each model sequentially to identify location mentions within the content. It ensures that the observed differences in output accurately reflect the inherent capabilities and limitations of each model.

### 4.2 Evaluation Metrics

In order to quantitatively assess the performance of the models in detecting geographical entities, our evaluation relies on three fundamental metrics: Precision, Recall, and F1 Score. Each metric provides a unique lens through which the effectiveness of the models can be scrutinized:

1. **Precision:** This metric measures the accuracy of the model in identifying positive instances. It quantifies the ratio of correctly identified geographical entities (True Positives) to the total number of entities the model identified (sum of True Positives and False Positives). High precision indicates that a model is effective in minimizing false positives. The formula for precision is:

    *Precision = (True Positives) / (True Positives + False Positives)*

2. **Recall:** This metric evaluates the model's ability to identify all relevant instances within the dataset. It is calculated as the ratio of True Positives to the total actual positives (sum of True Positives and False Negatives). High recall implies that the model effectively minimizes false negatives, capturing a higher proportion of actual geographical entities. The formula for recall is:

$$Recall = (True\ Positives) / (True\ Positives + False\ Negatives)$$

3. **F1 Score:** Serving as the harmonic mean of Precision and Recall, the F1 Score provides a single metric to assess the balance between Precision and Recall. It is particularly useful when the cost of false positives and false negatives varies or when one seeks a balance between identifying as many positives as possible while minimizing incorrect identifications. The F1 Score is defined as:

$$F1\ Score = 2 * (Precision * Recall) / (Precision + Recall)$$

To ensure the accuracy of these metrics, each geographical entity identified by the models was subjected to manual verification. This rigorous validation process allowed us to accurately categorize the results, providing a robust foundation for our calculations.

These metrics collectively offer a comprehensive view of each model's performance across the dataset, enabling us to identify strengths and weaknesses in the context of multilingual geo-entity detection.

### 4.3 Accuracy Assessment

The accuracy of each model was evaluated based on its ability to correctly identify geographical entities across the dataset. This assessment involved calculating the precision, recall, and F-1 score for each model, thereby providing a balanced view of each model's performance. These calculations highlight their strengths and areas for improvement in geo-entity detection.

### 4.4 Results Interpretation

The results from this evaluation offer invaluable insights into the comparative performance of advanced NLP models and LLMs in the domain of multilingual geo-entity detection. By analyzing the True Positives, False Positives, and False Negatives rates across different models and languages, we discern patterns of effectiveness, biases, and limitations inherent to each model. This detailed analysis not only benchmarks the current state-of-the-art but also paves the way for targeted improvements in NLP technologies for geospatial analysis.

In addition to a narrative interpretation, the inclusion of tables or graphs summarizing the evaluation results is planned, making it easier for readers to compare the performance metrics across models. The findings from this evaluation are anticipated to highlight potential areas for improvement in model accuracy, processing efficiency, and multilingual adaptability, guiding future research and development in NLP technologies for enhanced geospatial analysis.

## 5 ANALYSIS AND RESULTS

In this section, we delve into the comprehensive analysis of the performance of various NLP models and LLMs in recognizing location entities across English, Russian, and Arabic languages, as detailed in Table 1, to elucidate the nuanced capabilities and limitations revealed through our evaluation process.

Table 1: NER evaluation based on location entities for English, Russian and Arabic

| | English | | | Russian | | | Arabic | | |
|---|---|---|---|---|---|---|---|---|---|
| **Model** | Precision | Recall | F1 Score | Precision | Recall | F1 Score | Precision | Recall | F1 Score |
| SpaCy | **0.87** | 0.89 | 0.88 | 0.75 | 0.92 | 0.83 | 0.00 | 0.00 | 0.00 |
| XLM-RoBERTa | 0.83 | **0.98** | **0.90** | **0.85** | **0.97** | **0.91** | **0.78** | **0.93** | **0.84** |
| mLUKE | **0.87** | 0.87 | 0.87 | 0.84 | 0.87 | 0.86 | 0.00 | 0.00 | 0.00 |
| GeoLM | 0.54 | 0.96 | 0.69 | 0.22 | 0.11 | 0.15 | 0.00 | 0.00 | 0.00 |
| OpenAI's GPT 3.5 | 0.73 | 0.48 | 0.58 | 0.48 | 0.66 | 0.55 | 0.38 | 0.16 | 0.23 |
| OpenAI's GPT 4 | 0.86 | 0.94 | **0.90** | 0.85 | 0.96 | 0.90 | 0.77 | 0.71 | 0.74 |

<sup>a</sup> The best result for each metric and language is highlighted in bold, while the second-best result is indicated with an underline to facilitate easy comparison of the models' performance.

Our evaluation reveals insightful patterns regarding the performance of advanced NLP models and LLMs in detecting location entities across the three languages. Notably, while the precision of XLM-RoBERTa and OpenAI's GPT-4 isn't the highest in English, they both achieve the best F1 scores, indicating a balanced performance between precision and recall. This balance is critical in practical applications where both identifying as many relevant entities as possible (high recall) and ensuring the entities identified are correct (high precision) are crucial. mLUKE and SpaCy also demonstrate commendable performance in English, underscoring their effectiveness in handling location entities within this language.

In the Russian context, XLM-RoBERTa and GPT-4 continue to excel, showcasing their robustness and adaptability across different linguistic frameworks. mLUKE and SpaCy maintain good results, suggesting that their methodologies are somewhat effective in Russian language processing as well. However, it's important to highlight the contrast in performance for SpaCy when analyzing Arabic texts. The Multi-language Model of SpaCy, which significantly underperformed in Arabic, suggests a limitation in the model's ability to handle the linguistic complexities of Arabic. This limitation also adversely affects mLUKE, as it relies on SpaCy for initial span detection, indicating the importance of the underlying NER capabilities in multi-language models.

While GeoLM is designed to bridge the gap between NLP and geospatial sciences, our results indicate a notable performance disparity across languages, with significant drops in F1 scores for Russian and Arabic compared to English. This stark contrast not only highlights the model's limitations in cross-lingual transfer for geospatial entity recognition but also underscores the preliminary nature of our findings. Further research is essential to fully understand these multilingual capabilities and to enhance the model's effectiveness across a broader spectrum of languages.

It's worth to note that both GeoLM and mLUKE are constrained by a 512-token limit, which significantly impacts their ability to process and accurately identify location entities in longer texts, a limitation evident even in our dataset of brief Telegram posts that occasionally exceed this length.

OpenAI's GPT-3.5 presents a case of mediocre performance across the board. In separate tests, GPT-3.5 exhibited inconsistencies in detecting location entities, which we assess is the possible reason that led to lower overall scores. This inconsistency might be attributed to the model's general-purpose design, which, unlike models fine-tuned for specific NLP tasks or languages, may struggle with the nuanced requirements of geospatial entity recognition.

An interesting observation from our evaluation pertains to the models' handling of multi-word location entities, unconventional terminology, and offensive words. We encountered numerous instances where locations were composed of multiple words, such as "Beirut Rafic Hariri International Airport" or the Russian "Красное Шебекинского городского округа" (Krasnoe village, Shebekinsky District), posing a significant challenge for models to accurately recognize and classify these as single entities. Additionally, the detection of slang, made-up words, and offensive terms as locations further complicates the task. For example, the derogatory "Свинорейх" (a contemptuous term for Ukraine post-Euromaidan) and the Arabic phrase الكيان الصهيوني (commonly used to denigrate the Israeli state by referring to it as a "Zionist entity" rather than a country) highlight another layer of complexity. These instances likely stem from the models' reliance on tokenization and the presence (or absence) of specific terms within their training dictionaries, rather than a comprehensive understanding of context or the ability to discern between standard, pejorative, and offensive geographic references. These findings underscore the importance of advanced tokenization techniques, context-aware processing, and ethical considerations in enhancing the accuracy of geospatial entity recognition across diverse linguistic and cultural landscapes.

## 6  Conclusion

Our findings highlight several important considerations for future NLP and LLM research and applications. The superior F1 scores of XLM-RoBERTa and GPT-4 underscore the vital importance of achieving a balance between precision and recall. This balance is crucial for accurately identifying a broad spectrum of relevant entities while minimizing inaccuracies, which is indispensable in real-world cases across various contexts. Such scenarios include extracting locations from diverse internet sources as part of efforts in areas like national security, where the ability to quickly and accurately identify and analyze geospatial information can be pivotal. These capabilities allow for assessing threats, situation awareness, and informing decision-making processes in a timely and efficient manner.

The observed variability in model performance across languages underscores the inherent complexity of developing truly multilingual solutions. It signals an opportunity for substantial advancements through targeted research on cross-lingual training and architecture optimization. The potential to achieve even greater accuracy and recall by leveraging the complementary strengths of models like XLM-RoBERTa and GPT-4, whether through tandem use or enhanced coherence in their applications, represents a promising avenue for future work. Implementing such strategies could involve using XLM-RoBERTa for initial text understanding and entity recognition across languages, complemented by GPT-4 for tasks requiring advanced text generation, contextual nuance, or decision-making based on the recognized entities and context.

While integrating these models presents challenges, exploring strategies for effective synergy could revolutionize the development of more accurate, versatile, and culturally aware NLP and LLM systems. Additionally, the continuous improvement and testing of other models, including those like SpaCy's, GeoLM, and mLUKE, should not be overlooked. With the right enhancements and fine-tuning, these models have the potential to significantly contribute to the diversity and capability of NLP solutions, ensuring a comprehensive approach to addressing the wide range of linguistic and cultural challenges encountered in the field.

**ACKNOWLEDGMENTS**

This work was supported by the National Science Program "Security and Defense", which has received funding from the Ministry of Education and Science of the Republic of Bulgaria under the grant agreement no. Д01-74/19.05.2022.